# A Comprehensive Survey of Retrieval-Augmented Generation (RAG): Evolution, Current Landscape and Future Directions

*Shailja Gupta (Carnegie Mellon University, USA)*
*Rajesh Ranjan (Carnegie Mellon University, USA)*
*Surya Narayan Singh (BIT Sindri, India)*

## Abstract

This paper presents a comprehensive study of Retrieval-Augmented Generation (RAG), tracing its evolution from foundational concepts to the current state of the art. RAG combines retrieval mechanisms with generative language models to enhance the accuracy of outputs, addressing key limitations of LLMs. The study explores the basic architecture of RAG, focusing on how retrieval and generation are integrated to handle knowledge-intensive tasks. A detailed review of the significant technological advancements in RAG is provided, including key innovations in retrieval-augmented language models and applications across various domains such as question-answering, summarization, and knowledge-based tasks. Recent research breakthroughs are discussed, highlighting novel methods for improving retrieval efficiency. Furthermore, the paper examines ongoing challenges such as scalability, bias, and ethical concerns in deployment. Future research directions are proposed, with a focus on improving the robustness of RAG models, expanding the scope of application of RAG models, and addressing societal implications. This survey aims to serve as a foundational resource for researchers and practitioners in understanding the potential of RAG and its trajectory in the field of natural language processing.

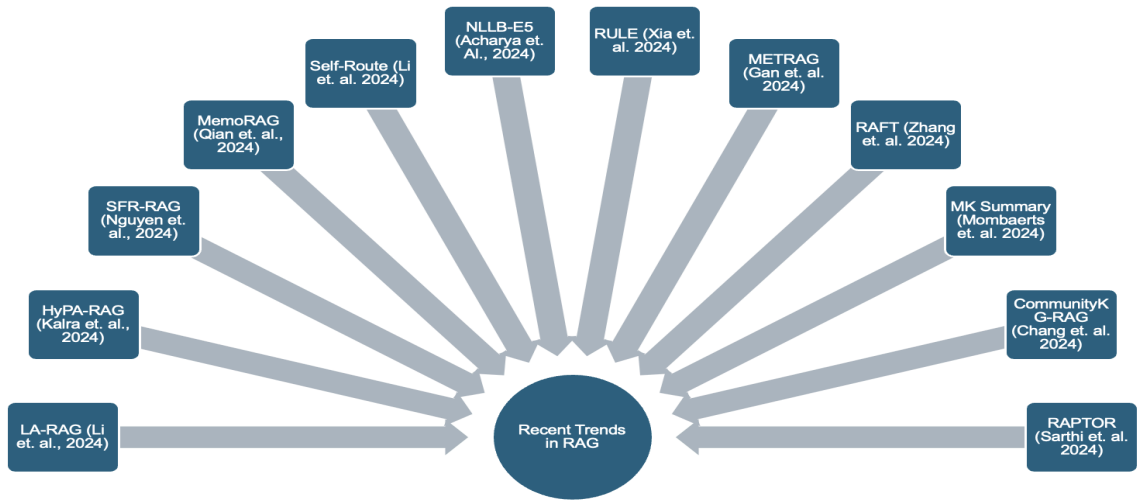

Figure 1: Trends in RAG captured from recent research papers

**Introduction**

**1.1 Introduction of Natural Language Generation (NLG)**

Natural Language Processing (NLP) has become a pivotal domain within artificial intelligence (AI), with applications ranging from simple text classification to more complex tasks such as summarization, machine translation, and question answering. A particularly significant branch of NLP is Natural Language Generation (NLG), which focuses on the production of human-like language from structured or unstructured data. NLG's goal is to enable machines to generate coherent, relevant, and context-aware text, improving interactions between humans and machines (Gatt et. al. 2018). As AI evolves, the demand for more contextually aware and factually grounded generated content has increased, bringing about new challenges and innovations in NLG.

Traditional NLG models, especially sequence-to-sequence architectures (Sutskever et al. 2014), have exhibited significant advancements in generating fluent and coherent text. However, these models tend to rely heavily on training data, often struggling when tasked with generating factually accurate or contextually rich content for queries that require knowledge beyond their training set. As a result, models like GPT (Radford et al. 2019) or BERT-based (Devlin et al. 2019) text generators are prone to hallucinations, where they produce plausible but incorrect or non-existent information (Ji et al. 2022). This limitation has prompted the exploration of hybrid models that combine retrieval mechanisms with generative capabilities to ensure both fluency and factual correctness in outputs. There has been a significant rise in several research papers in this field and several new methods across the RAG components have been proposed. Apart from new algorithms and methods, RAG has also seen steep adoption across various applications. However, there is a gap in a sufficient survey of this space tracking the evolution and recent changes in this space. The current survey intends to fill this gap.

**1.2 Overview of Retrieval-Augmented Generation (RAG)**

Retrieval-Augmented Generation (RAG) is an emerging hybrid architecture designed to address the limitations of pure generative models. RAG integrates two key components: (i) a retrieval mechanism, which retrieves relevant documents or information from an external knowledge source, and (ii) a generation module, which processes this information to generate human-like text (Lewis et al. 2020). This combination allows RAG models to not only generate fluent text but also ground their outputs in real-world, up-to-date data.

The retrieval module in RAG typically leverages dense vector representations to identify relevant documents from large datasets, such as Wikipedia or proprietary databases. Once retrieved, these documents are passed to the generative module, often built using transformer-based architectures, to generate responses grounded in the retrieved knowledge. This methodology helps mitigate the hallucination problem and ensures that the generated text is more factual and contextually appropriate (Thakur et al. 2021). Over the period, RAG models have seen applications in various domains, including open-domain question answering (Karpukhin et al., 2020), conversational agents (Liu et al. 2021), and personalized recommendations.

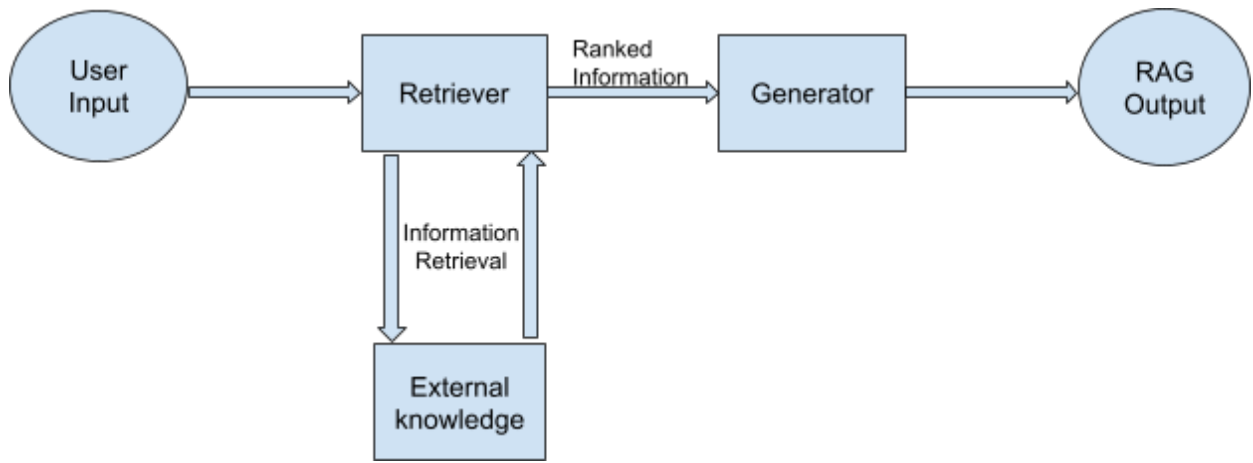

Figure 2: A basic flow of the RAG system along with its component

**1.3 Evolution of Hybrid Models in NLP**

Before the introduction of RAG, NLP models primarily relied on either retrieval or generation approaches, each with its own set of advantages and limitations. Retrieval-based systems, such as traditional information retrieval engines (Salton et al., 1975), efficiently provided relevant documents or snippets in response to a query but could not synthesize new information or present the results in a coherent narrative. On the other hand, purely generative models, which became popular with the rise of transformer architectures (Vaswani et al. 2017), offered fluency and creativity but often lacked factual accuracy.

The development of hybrid systems combining retrieval and generation began to gain momentum as researchers recognized the complementary strengths of both approaches. Early efforts in hybrid modeling can be traced back to works like DrQA (Chen et al. 2017), which employed retrieval techniques to fetch relevant documents for question-answering tasks. However, the generative component in such systems was minimal, often limited to selecting text directly from the retrieved documents. Similarly, in models like Information Retrieval (Dai et al. 2019), retrieval was treated as distinct, independent components.

The real innovation came with the realization that retrieval and generation could be tightly integrated. Models like REALM (Guu et al., 2020) represented a key milestone, as they trained the retrieval and generative components jointly, enabling better alignment between the retrieved information and the generated output. RAG (Lewis et al. 2020) further extended this paradigm by using dense passage retrieval (Karpukhin et al., 2020) to fetch relevant documents and transformers like BART (Lewis et al., 2020) for a generation. This architecture provided a more seamless integration of retrieval and generation, allowing the model to answer open-ended questions with both fluency and factual grounding.

**1.4 Importance of Factually Grounded Language Generation**

One of the main motivations for developing RAG is the increasing demand for factually accurate, contextually relevant, and up-to-date generated content. In many applications, such as customer service, medical diagnostics, or legal advisory systems, the need for reliable and grounded responses is paramount. Generative models that produce hallucinated or inaccurate information can lead to serious consequences, such as spreading misinformation or providing incorrect advice (Ji et al. 2022).

RAG models directly address these concerns by grounding their generative process in external, up-to-date knowledge sources. This grounding improves the factual accuracy of the output and enhances the relevance of responses by incorporating real-world data that is directly tied to the query. Additionally, RAG models are less likely to propagate biases present in static training data, as they can retrieve more diverse and balanced information from external sources

## 1.5 Applications of RAG Models

RAG models have been applied across a wide array of domains where factual accuracy and contextual understanding are critical. One of the most prominent applications is in open-domain question answering, where the model must generate answers based on a wide range of topics. RAG has proven effective in improving answer accuracy by retrieving relevant information and then generating responses grounded in that data (Izacard et. al. 2021). Models like Dense Passage Retrieval (DPR) (Karpukhin et al., 2020) and Fusion-in-Decoder (Izacard et. al. 2021) have been used to great effect in this context, showing significant improvements over traditional generative or retrieval-only models.

In conversational AI, RAG models have enhanced the capabilities of dialogue systems by ensuring that responses are both coherent and grounded in factual information (Roller et al., 2020). For example, chatbots used in customer service can benefit from RAG's ability to retrieve specific details from product databases or documentation, leading to more accurate and useful responses for end-users.

Other applications include medical diagnosis systems, where RAG can retrieve and integrate the latest research findings or patient-specific data to generate accurate diagnostic suggestions, and legal advisory systems, where the model can retrieve relevant case law or statutes to provide legally sound advice. Furthermore, RAG has found applications in personalized recommendation systems, where it can retrieve user preferences or past interactions and generate personalized suggestions.

## 1.6 Challenges and Limitations of RAG

Despite the promise of RAG models, several challenges need attention. The retrieval mechanism, while powerful, can still struggle with retrieving the most relevant documents, particularly when dealing with ambiguous queries or niche knowledge domains. The reliance on dense vector representations, such as those used in DPR, can sometimes lead to irrelevant or off-topic documents being retrieved. Efforts to refine retrieval techniques, including the incorporation of more sophisticated query expansion and contextual disambiguation, are needed to improve performance in these areas. The integration between retrieval and generation, while seamless in theory, can sometimes fail in practice. For instance, the generative module may not always effectively incorporate the retrieved information into its responses, leading to inconsistencies or incoherence between the retrieved facts and the generated text. Research into better alignment mechanisms, such as improved attention models or hierarchical fusion techniques, may help alleviate these issues (Izacard et. al. 2021). Additionally, the computational overhead of RAG models is a concern, as they require both a retrieval and a generation step for each query. This dual process can be resource-intensive, particularly for large-scale applications (Borgeaud et al. 2021). Techniques such as model pruning (Han et al. 2015) or knowledge distillation (Sanh et al., 2019) may offer ways to reduce the computational burden without sacrificing performance. Finally, there are ethical concerns associated with the deployment of RAG models, particularly in terms of bias and transparency. Biases in AI and LLM have been a well-researched and evolving field with researchers identifying different types of biases not limited to Gender, socio-economic class, or even educational background (Gupta et. al. 2024; Ranjan et. al., 2024). While RAG has the potential to reduce biases by retrieving more balanced information, there is still the risk of amplifying biases present in the retrieved sources

(Binns, 2018). Furthermore, ensuring transparency in how retrieval results are selected and used in generation is crucial for maintaining trust in these systems.

### 1.7 Scope of the Survey

This paper aims to provide a comprehensive survey of RAG models, covering their evolution, key architectural components, recent research in this area, current challenges and limitations of RAG, and future research direction.

## 2: Core Components and Architectural Overview of RAG Systems

### 2.1 Overview of RAG Models

Retrieval-augmented generation (RAG) is an advanced hybrid model architecture that augments natural language generation (NLG) with external retrieval mechanisms to enhance the model's knowledge base. Traditional large language models (LLMs) such as GPT-3 and BERT, which are pre-trained on vast corpora, rely entirely on their internal representations of knowledge, making them susceptible to issues like hallucinations—where the models generate plausible but incorrect information. These models cannot efficiently update their knowledge bases without retraining, making them less practical for dynamic, knowledge-intensive tasks like open-domain question answering and fact verification (Brown, T., et al. 2020). To overcome these limitations, the paper (Lewis et al. 2020) proposed the RAG architecture, which retrieves real-time, relevant external documents to ground the generated text in factual information.

The RAG model incorporates two key components:

1. **Retriever**: This retrieves the most relevant documents from a corpus using techniques such as dense passage retrieval (DPR) (Karpukhin et. al. *2020*) or traditional BM25 algorithms.
2. **Generator**: It synthesizes the retrieved documents into coherent, contextually relevant responses.

RAG's strength lies in its ability to leverage external knowledge dynamically, allowing it to outperform generative models like GPT-3 and knowledge-grounded systems like BERT, which rely on static datasets. In open-domain question answering, RAG has been demonstrated to be highly effective, consistently retrieving relevant information and improving the factual accuracy of the generated responses (Guu, K., et al. 2020). In addition to knowledge retrieval, RAG models excel at updating knowledge bases. Since the model fetches external documents for each query, it requires no retraining to incorporate the latest information. This flexibility makes RAG models particularly suitable for domains where information is constantly evolving, such as medical research, financial news, and legal proceedings. Furthermore, studies have shown that RAG models achieve superior results in a variety of knowledge-intensive tasks, including document summarization and, knowledge-grounded dialogues

### 2.2 Retriever Mechanisms in RAG Systems

The retriever in RAG systems is essential for fetching relevant documents from an external corpus. Effective retrieval ensures that the model's output is grounded in accurate information. Several retrieval mechanisms are commonly used, ranging from traditional methods like BM25 to more sophisticated techniques like Dense Passage Retrieval (DPR).

**2.2.1 BM25**

BM25 is a well-established information retrieval algorithm that uses the term frequency-inverse document frequency (TF-IDF) to rank documents according to relevance. Despite being a classical method, BM25 remains a strong baseline for many modern retrieval systems, including those used in RAG models. BM25 calculates the relevance score of a document based on how frequently a query term appears in the document while adjusting for the document's length and the frequency of the term across the corpus (Robertson et. al. 2009). While BM25 is effective for keyword matching, it has limitations in understanding semantic meaning. For example, BM25 cannot capture the relationships between words and tends to perform poorly on more complex, natural language queries that require an understanding of context. Despite this limitation, BM25 is still widely used because of its simplicity and efficiency. BM25 is effective for tasks involving simpler, keyword-based queries, although more modern retrieval models like DPR tend to outperform it in semantically complex tasks.

**2.2.2 Dense Passage Retrieval (DPR)**

Dense Passage Retrieval (DPR), introduced by Karpukhin et al. (2020), represents a more modern approach to information retrieval. It uses a dense vector space in which both the query and the documents are encoded into high-dimensional vectors. DPR employs a bi-encoder architecture, where the query and documents are encoded separately, allowing for efficient nearest-neighbor search (Xiong et. al. 2020). Unlike BM25, DPR excels at capturing semantic similarity between the query and documents, making it highly effective for open-domain question-answering tasks. The strength of DPR lies in its ability to retrieve relevant information based on semantic meaning rather than keyword matching. By training the retriever on a large corpus of question-answer pairs, DPR can find documents that are contextually related to the query, even when the query and the document do not share exact terms. Recent research has further improved DPR by integrating it with pre-trained language models and an example is LLM adapted for the dense RetrievAl approach (Li et. al. 2023)

**2.2.3 REALM (Retrieval-Augmented Language Model)**

Another significant advancement in retrieval mechanisms for RAG models is REALM (Guu et al. (2020). REALM integrates retrieval into the language model's pre-training process, ensuring that the retriever is optimized alongside the generator for downstream tasks. The key innovation in REALM is that it learns to retrieve documents that improve the model's performance on specific tasks, such as question answering or document summarization. During training, REALM updates both the retriever and the generator, ensuring that the retrieval process is optimized for the generation task. REALM's retriever is trained to identify documents that are not only relevant to the query but also helpful for generating accurate and coherent responses. As a result, REALM significantly improves the quality of generated responses, particularly in tasks that require external knowledge. Recent studies have demonstrated that REALM outperforms both BM25 and DPR in certain knowledge-intensive tasks, particularly when retrieval is tightly coupled with generation.

The core of RAG lies in the quality of retrieved passages, but many current methods rely on similarity-based retrieval (Mallen et al. 2022). Self-RAG (Asai et al. 2023b), and REPLUG (Shi et al., 2023) have advanced by leveraging LLMs to enhance retrieval capabilities, achieving more adaptive retrieval. After initial retrieval, cross-encoder models are used to re-rank the retrieved results by jointly encoding the query and each retrieved document to compute relevance scores. These models provide more context-aware retrieval at the cost of higher computational overhead. Pointwise and Pairwise Ranking, often based on Learning-to-Rank (LTR) algorithms, are used to assign relevance scores to retrieved documents, either independently (pointwise) or by comparing document pairs (pairwise). RAG systems utilize self-attention within the LLM to manage context and relevance across different parts of the input and retrieved text. Cross-attention mechanisms are used when integrating retrieved information into the generative model, ensuring that the most relevant pieces of information are emphasized during generation.

**2.3 Generator Mechanisms in RAG Systems**

In Retrieval-Augmented Generation (RAG) systems, the generator mechanism plays a crucial role in producing the final output by integrating retrieved information with the input query. After the retrieval component pulls relevant knowledge from external sources, the generator synthesizes this information into coherent, contextually appropriate responses. The Large Language Model (LLM) serves as the backbone of the generator, which ensures the generated text is fluent, accurate, and aligned with the original query.

**2.3.1 T5 (Text-to-Text Transfer Transformer)**

T5 (Text-to-Text Transfer Transformer) (Raffel et al. 2020) is one of the most commonly used models for generation tasks in RAG systems. T5 is versatile in its approach, framing every NLP task as a text-to-text task. This uniform framework allows T5 to be fine-tuned for a wide range of tasks, including question-answering, summarization, and dialogue generation. By integrating retrieval with generation, T5-based RAG models have been shown to outperform traditional generative models like GPT-3 and BART on several benchmarks, including the Natural Questions dataset and the TriviaQA dataset. Moreover, T5's ability to handle complex multi-task learning makes it a popular choice for RAG systems that need to tackle a diverse range of knowledge-intensive tasks.

**2.3.2 BART**

BART (Bidirectional and Auto-Regressive Transformer), introduced by Lewis et al. (2020), is another prominent generative model used in RAG systems. BART is particularly well-suited for tasks involving text generation from noisy inputs, such as summarization and open-domain question answering. As a denoising autoencoder, BART can reconstruct corrupted text sequences, making it robust for tasks that require the generation of coherent, factual outputs from incomplete or noisy data. When paired with a retriever in a RAG system, BART has been shown to improve the factual accuracy of generated text by grounding it in external knowledge. Studies have demonstrated that BART-based RAG models achieve state-of-the-art results in various knowledge-intensive tasks, including dialogue generation and news summarization.

**3. Retrieval-Augmented Generation Models Across Different Modalities**

**3.1 Text-Based RAG Models:** Text-based RAG models represent the most mature and widely researched category. These models leverage textual data for both retrieval and generation tasks, enabling applications such as question-answering, summarization, and conversational agents. Transformer architectures, such as BERT (Devlin et al., 2019) and T5 (Raffel et al., 2020), are foundational in text-based RAG models. These models utilize self-attention mechanisms to capture contextual relationships within text, which enhances both retrieval accuracy and generation fluency. Dense retrieval models, such as those using dense embeddings from BERT, offer superior performance compared to traditional sparse methods like TF-IDF. Dense retrievers (Karpukhin et al. 2020), leverage dense representations to retrieve relevant documents more effectively. Recent advancements focus on integrating retrieval and generation into a single training pipeline. REALM (Guu et al., 2020) is an example of such an end-to-end model that jointly optimizes retrieval and generation processes, improving overall task performance.

**3.2 Audio-Based RAG Models:** Audio-based RAG models extend the principles of retrieval-augmented generation to the audio modality, enabling applications such as speech recognition, audio summarization, and conversational agents in voice interfaces. Audio data is often represented using embeddings derived from pre-trained models like Wav2Vec 2.0 (Baevski et al., 2020). These embeddings serve as input to retrieval and generation components, enabling the model to handle audio data effectively.

**3.3 Video-Based RAG Models:** Video-based RAG models incorporate both visual and textual information to enhance performance in tasks such as video understanding, captioning, and retrieval. Video data is represented using embeddings from models like I3D (Xie et. al. 2017) or TimeSformer (Bertasius et al. 2021). These embeddings capture temporal and spatial features essential for effective retrieval and generation.

**3.4 Multimodal RAG Models:** Multimodal RAG models integrate data from multiple modalities—text, audio, video, and images—to provide a more holistic approach to retrieval and generation tasks. Models like Flamingo (Alayrac et al., 2022) integrate multiple modalities into a unified framework, enabling simultaneous processing of text, images, and videos. Techniques for cross-modal retrieval involve retrieving relevant information across different modalities (Li. et. al. 2023).

Multimodal capabilities enhance the versatility and efficiency of RAG across various applications." Retrieval as generation" (Wang et. al. 2024) extends the Retrieval-Augmented Generation (RAG) framework to multimodal applications by incorporating text-to-image and image-to-text retrieval. Utilizing a large dataset of paired images and text descriptions, the system accelerates image generation when user queries align with stored text descriptions ("retrieval as generation"). The image-to-text functionality allows users to engage in discussions based on input images.

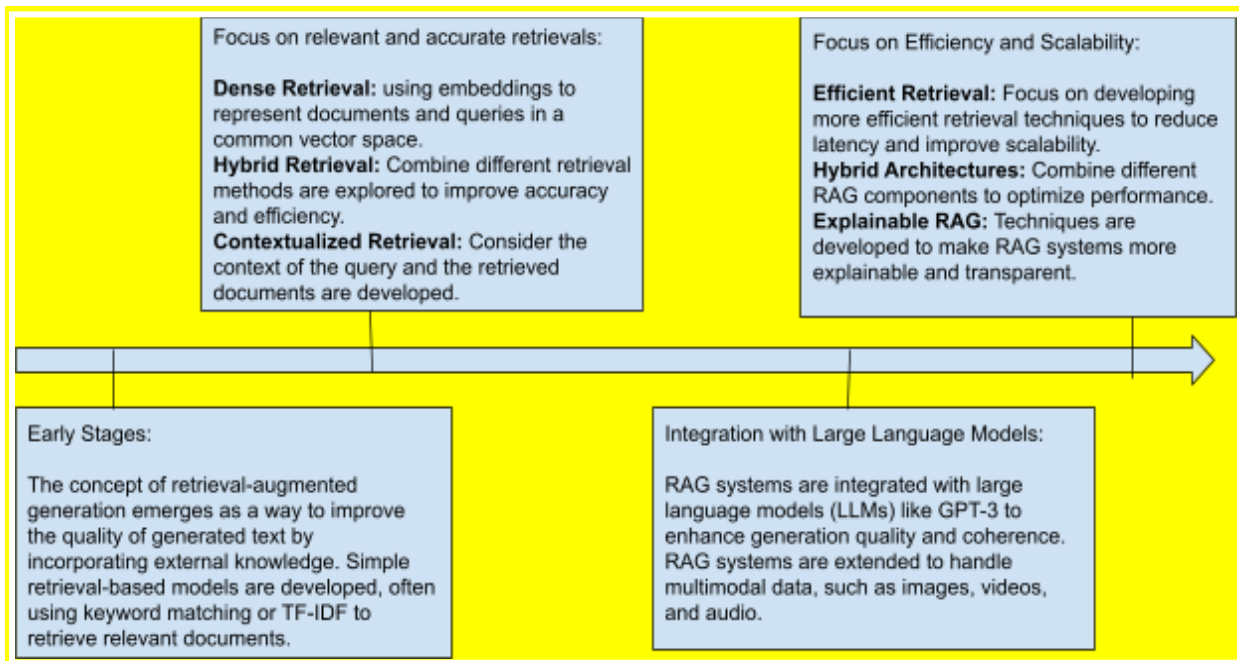

Figure 3: Timeline of the evolution of the RAG system and its components

## 4. Recent Advancement in the field:

There has been significant advancement in this field and this section intends to capture key findings of a few important recent papers. A novel agentic Retrieval-Augmented Generation (RAG) framework (Ravuru et. al. 2024) employs a hierarchical, multi-agent architecture where specialized sub-agents, using smaller pre-trained language models (SLMs), are fine-tuned for specific time series tasks. The master agent delegates tasks to these sub-agents, who retrieve relevant prompts from a shared knowledge repository. In this modular, multi-agent approach, the authors achieve state-of-the-art performance demonstrating improved flexibility and effectiveness over task-specific methods in time series analysis. RULE (Xia et. al. 2024), a multimodal Retrieval-Augmented Generation (RAG) framework designed to improve the factuality of medical Vision-Language Models (Med-LVLM), addresses challenges in medical RAG by introducing a calibrated selection strategy to control factuality risk, and, by developing a preference optimization strategy to balance the model's intrinsic knowledge with retrieved contexts, proving its effectiveness in enhancing factual accuracy in Med-LVLM systems. METRAG (Gan et. al. 2024), a multi-layered, thoughts-enhanced retrieval-augmented generation framework, integrates LLM supervision to generate utility-oriented thoughts and combines document similarity with utility for improved performance. It also incorporates a task-adaptive summarizer to produce compact thoughts. Using the multi-layered thoughts from these stages, an LLM generates knowledge-augmented content, demonstrating superior performance on knowledge-intensive tasks compared to traditional approaches. Distractor document is one of the key traits of Retrieval Augmented Fine-Tuning (RAFT) (Zhang et. al. 2024) where the model is trained to disregard irrelevant, distractor documents and instead cite directly from relevant sources. This process, combined with a chain-of-thought reasoning style, enhances the model's reasoning capabilities. RAFT demonstrates consistent performance improvements in domain-specific RAG tasks, including PubMed, HotpotQA, and Gorilla datasets, serving as a post-training enhancement for LLMs. FILCO (Wang et. al. 2023), a method designed to enhance the quality of context provided to generative models in tasks like open-domain question answering and fact verification, addresses issues of over- or under-reliance on retrieved passages, which can lead to problems such as hallucinations in the generated outputs. The method improves context quality by identifying useful context through lexical and information-theoretic approaches and training context filtering models to refine retrieved contexts during test time. Reflection Token is a key attribute of Self-reflective Retrieval Augmented-Generation (Self-RAG) (Asai et. al. 2023), a novel framework designed to improve the factual accuracy of large language models (LLMs) by combining retrieval with self-reflection. Unlike traditional methods that retrieve and incorporate a fixed number of passages, Self-RAG adaptively retrieves relevant passages and uses reflection tokens to evaluate and refine its responses, allowing the model to adjust its behavior according to task-specific needs and has shown superior performance in open-domain question-answering, reasoning, fact verification, and long-form generation tasks. Intelligence and effectiveness of RAG are dependent a lot on the quality of retrieval and more meta-data understanding of the repository would enhance the effectiveness of the RAG system. A novel data-centric Retrieval-Augmented Generation (RAG) workflow advances beyond the traditional retrieve-then-read mode and employs a prepare-then-rewrite-then-retrieve-then-read framework, enhancing LLMs by integrating contextually relevant, time-critical, or domain-specific information. Key innovations include generating metadata, synthetic Questions and Answers (QA), and introducing the Meta Knowledge Summary (MK Summary) for clusters of documents (Mombaerts et. al. 2024). A recent paper introduces CommunityKG-RAG (Chang et. al. 2024), a zero-shot framework that integrates community structures within Knowledge Graphs (KGs) into Retrieval-Augmented Generation (RAG) systems. This approach enhances the accuracy and contextual relevance of fact-checking by utilizing multi-hop connections within KGs, outperforming traditional methods without requiring additional domain-specific training. The RAPTOR model (Sarthi et. al. 2024) introduces a hierarchical approach to retrieval-augmented language models, addressing limitations in traditional methods that retrieve only short, contiguous text chunks. RAPTOR forms a summary tree to retrieve information at varying abstraction levels by recursively embedding, clustering, and summarizing text. Experiments demonstrate RAPTOR's superior performance, especially in question-answering tasks requiring complex reasoning. When paired with GPT-4, RAPTOR improves accuracy on the QuALITY benchmark by 20%.

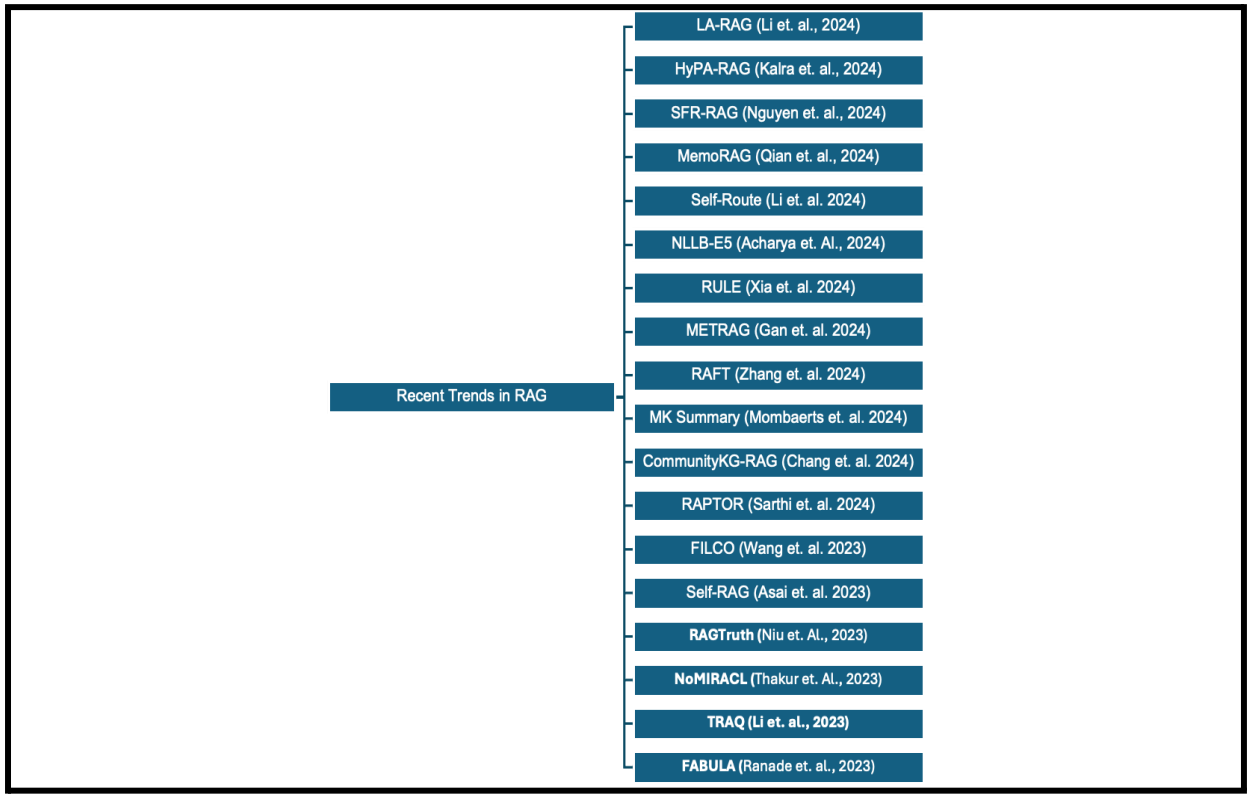

Figure 4: Evolving Trends in RAG captured from research papers

This advancement in RAG further proves the utility of the RAG system however recent LLM launches that support long-term context have significantly shown improved performance. A recent study (Li et. al. 2024) compared the efficiency of Retrieval Augmented Generation (RAG) and long-context (LC) Large Language Models (LLMs), such as Gemini-1.5 and GPT-4. While LC models outperform RAG when adequately resourced, RAG's cost-efficiency remains advantageous. To balance performance and cost, the paper introduces Self-Route. This method dynamically directs queries to either RAG or LC based on model self-reflection, optimizing both computation cost and performance. This study offers valuable insights into the optimal application of RAG and LC in handling long-context tasks. Nguyen et. al., 2024 introduce SFR-RAG, a small but highly efficient Retrieval Augmented Generation (RAG) model, which is designed to enhance the integration of external contextual information into Large Language Models (LLMs) while minimizing hallucinations. LA-RAG (Li et. al., 2024), a novel Retrieval-Augmented Generation (RAG) paradigm designed to enhance Automatic Speech Recognition (ASR) in large language models (LLMs). One of the key benefits of LA-RAG is its ability to leverage fine-grained token-level speech data stores alongside a speech-to-speech retrieval mechanism, improving ASR accuracy by incorporating LLM in-context learning (ICL). The study focuses on datasets of Mandarin and various Chinese dialects, demonstrating significant accuracy improvements, particularly in managing accent variations, which have historically been a challenge for existing speech encoders. The findings highlight LA-RAG's potential to advance ASR technology, offering a more robust solution for diverse acoustic conditions. Large Language Models (LLMs) face challenges in AI legal and policy contexts due to outdated knowledge and hallucinations. HyPA-RAG (Kalra et. al., 2024), a Hybrid Parameter-Adaptive Retrieval-Augmented Generation system, improves accuracy by using adaptive parameter tuning and hybrid retrieval strategies. Tested on NYC Local Law 144 (LL144), HyPA-RAG demonstrates enhanced correctness and contextual precision, addressing the complexities of legal texts. MemoRAG (Qian et. al., 2024) introduces a novel Retrieval-Augmented Generation (RAG) paradigm designed to overcome the limitations of traditional RAG systems in handling ambiguous or unstructured knowledge. MemoRAG's dual-system architecture utilizes a lightweight long-range LLM to generate draft answers and guide retrieval tools, while a more powerful LLM refines the final output. This framework, optimized for better cluing and memory capacity, significantly outperforms conventional RAG models across both complex and straightforward tasks. NLLB-E5 (Acharya et. al., 2024) introduces a scalable multilingual retrieval model aimed at addressing the challenges faced in supporting multiple languages, particularly low-resource languages like Indic languages. By leveraging the NLLB encoder and a distillation approach from the E5 multilingual retriever, NLLB-E5 enables zero-shot retrieval across languages without the need for multilingual training data. Evaluations on benchmarks such as Hindi-BEIR showcase its robust performance, highlighting task-specific challenges and advancing multilingual information access for global inclusivity.

## 5. Current Challenges and Limitations in Retrieval-Augmented Generation (RAG):

This section intends to highlight the current challenges and limitations of RAG considering the current landscape of the system and this would shape the future research directions in the field.

**Scalability and Efficiency:** One of the primary challenges for RAG models is scalability. As retrieval components rely on external databases, handling vast and dynamically growing datasets requires efficient retrieval algorithms. High computational costs and memory requirements also make it difficult to deploy RAG models in real-time or resource-constrained environments (Shi et al. 2023), (Asai et al. 2023b).

**Retrieval Quality and Relevance:** Ensuring the quality and relevance of retrieved documents remains a significant concern. Retrieval models can sometimes return irrelevant or outdated information, which negatively affects the accuracy of the generated output. Improving retrieval precision, especially for long-form content generation, remains an active area of research (Mallen et al. 2022), (Shi et al. 2023).

**Bias and Fairness:** Similar to other machine learning models, RAG systems can exhibit bias due to biases present in the retrieved datasets. Retrieval-based models may amplify harmful biases in retrieved knowledge, leading to biased outputs in a generation. Developing bias mitigation techniques for retrieval and generation in tandem is an ongoing challenge.

**Coherence:** RAG models often struggle with integrating the retrieved knowledge into coherent, contextually relevant text. The alignment between retrieved passages and the generation model's output is not always seamless, leading to inconsistencies or factual hallucinations in the final response (Ji et al. 2022).

**Interpretability and Transparency:** Like many AI systems, RAG models are often treated as black boxes, with limited transparency in how retrieval influences generation. Improving the interpretability of these models is crucial to fostering trust, especially in critical applications (Roller et al. 2020).

## 6. Future Research Directions for Retrieval-Augmented Generation (RAG)

Retrieval-augmented generation (RAG) represents a significant advancement in natural language processing and related fields by combining retrieval and generative mechanisms. This section explores key areas for future research, highlighting the potential for innovation and improvement in RAG systems.

**6.1 Enhancing Multimodal Integration:** The integration of text, image, audio, and video data in RAG models remains an evolving challenge. Future research should focus on improving multimodal fusion techniques to enable seamless interaction between different data types. This includes developing advanced methods for aligning and synthesizing information across modalities. Recent works (Chen et. al. 2022), (Yasunaga et. al. 2022), (Zhu et. al. 2024) have explored multimodal learning, but further innovations are needed to enhance the coherence and contextuality of multimodal outputs.Research into cross-modal retrieval aims to improve the ability of RAG systems to retrieve relevant information across different modalities. For example, combining text-based queries with image or video content retrieval could enhance applications such as visual question answering and multimedia search. **This is another future direction to explore for RAG related research.**

**6.2 Scaling and Efficiency:** As RAG models are deployed in increasingly large-scale applications, scalability becomes a critical concern. Research should focus on developing methods to efficiently scale retrieval and generation processes without compromising performance. Techniques such as distributed computing and efficient indexing methods are essential for handling large datasets. Improving the efficiency of RAG models involves optimizing both retrieval and generation components to reduce computational resources and latency.

**6.3 Personalization and Adaptation:** Future RAG models should focus on personalizing retrieval processes to cater to individual user preferences and contexts. This involves developing techniques to adapt retrieval strategies based on user history, behaviour, and preferences. Enhancing the contextual adaptation of RAG models by deeper understanding of the context and sentiments of query (Gupta et. al. 2024) and the repository of ducments is crucial for improving the relevance of generated responses. Research should explore methods for dynamic adjustment of retrieval and generation processes based on the evolving context of interactions. This includes incorporating user feedback and contextual cues into the RAG pipeline.

**6.4 Ethical and Privacy Considerations:** Addressing biases (Shrestha et. al. 2024), (Gupta et. al. 2024) in general and specifics to RAG models is a critical area for future research. As RAG systems are deployed in diverse applications, ensuring fairness and mitigating biases in retrieved and generated content is essential. Future RAG research should focus on privacy-preserving techniques to protect sensitive information during retrieval and generation. This includes developing methods for secure data handling and privacy-aware retrieval strategies. Interpretability of model is also a critical area to focus upon as a part of on going research in improving RAG.

**6.5 Cross-Lingual and Low-Resource Languages:** Expanding RAG technology to support multiple languages ( Chirkova et. al. 2024), especially low-resource languages, is a promising direction. Future research should aim to improve cross-lingual retrieval and generation capabilities to provide accurate and relevant results across different languages. Enhancing RAG models to effectively support low-resource languages involves developing methods to retrieve and generate content with limited training data. Research should focus on techniques for transfer learning and data augmentation to improve performance in underrepresented languages.

**6.6 Advanced Retrieval Mechanisms:** Future RAG research should explore dynamic retrieval mechanisms that adapt to changing query patterns and content requirements. This includes developing models that can dynamically update their retrieval strategies based on new information and evolving user needs. Investigating hybrid retrieval approaches that combine various retrieval strategies, such as dense and sparse retrieval, could enhance the effectiveness of RAG systems. Research should explore how to integrate different retrieval methods to achieve optimal performance for diverse tasks.

**6.7 Integration with Emerging Technologies:** Integrating RAG models with brain-computer interfaces (BCIs) could lead to novel applications in human-computer interaction and assistive technologies. Research should explore how RAG systems can leverage BCI data to enhance user experience and generate context-aware responses.The integration of RAG with AR and VR technologies presents opportunities for creating immersive and interactive experiences. Future research should investigate how RAG models can be used to enhance AR and VR applications by providing contextually relevant information and interactions.

## 7. Conclusion

Retrieval-Augmented Generation (RAG) has undergone significant evolution, with extensive research dedicated to improving retrieval effectiveness and enhancing coherent generation to minimize hallucinations. From its early iterations to recent advancements, RAG has been instrumental in integrating external knowledge into Large Language Models (LLMs), thereby boosting accuracy and reliability. In particular, recent domain-specific work has showcased RAG's potential in specialized areas such as legal, medical, and low-resource language applications, highlighting its adaptability and scope. However, despite these advances, this paper identifies clear gaps that remain unresolved. Challenges such as the integration of ambiguous or unstructured information, effective handling of domain-specific contexts, and the high computational overhead of complex retrieval tasks still persist. These limitations constrain the broader applicability of RAG systems, particularly in diverse and dynamic real-world environments. The future research directions outlined in this paper—ranging from improving retrieval mechanisms to enhancing context management and ensuring scalability—will serve as a critical guide for the next phase of innovation in this space. By addressing these gaps, the next generation of RAG models has the potential to drive more reliable, efficient, and domain-adaptable LLM systems, further pushing the boundaries of what is possible in retrieval-augmented AI applications.